\pdfoutput=1
\documentclass[11pt]{article}

\usepackage{ACL2023}

\usepackage{times}
\usepackage{multirow}

\usepackage{latexsym}
\usepackage{booktabs}

\usepackage{graphicx}
\graphicspath{ {./img/} }

\usepackage[T1]{fontenc}
\usepackage[utf8]{inputenc}

\usepackage{microtype}

\usepackage{inconsolata}

\title{HULAT at SemEval-2023 Task 9: Data augmentation for pre-trained transformers applied to Multilingual Tweet Intimacy Analysis}
\author{Isabel Segura-Bedmar \\
  Computer Science Department, Universidad Carlos III de Madrid, Leganés, Spain \\ \texttt{isegura@inf.uc3m.es} \\}

\begin{document}
\maketitle
\begin{abstract}

This paper describes our participation in SemEval-2023 Task 9, Intimacy Analysis of Multilingual Tweets. We fine-tune some of the most popular transformer models with the training dataset and synthetic data generated by different data augmentation techniques. During the development phase, our best results were obtained by using XLM-T. Data augmentation techniques provide a very slight improvement in the results. Our system ranked in the 27th position out of the 45 participating systems. Despite its modest results, our system shows promising results in languages such as Portuguese, English, and Dutch. All our code is available in the repository \url{https://github.com/isegura/hulat_intimacy}. 

\end{abstract}

\section{Introduction}
The Intimacy Analysis of Multilingual Tweets \cite{pei2022semeval} is a very novel task whose goal is to estimate the level of intimacy of a text. If we can detect the degree of intimacy of a text, this could help us better understand the social norms that exist within a culture or society. Furthermore, from a computational point of view, this task will allow us to assess whether current computational models are capable of identifying the intimacy level of a text, distinguishing between texts with very intimate content and texts without any intimate information. 

Although the automatic analysis of textual intimacy in language could  be very beneficial for politicians, governments, sociologists, anthropologists, companies, and so on, very few research efforts have been devoted to this challenging task to now \cite{pei-jurgens-2020-quantifying}. Another important aspect of this task is its multilingual setting. The dataset used in the task is the MINT dataset \cite{pei-jurgens-2020-quantifying} with a total of 13,384 tweets in 10 languages including English, French, Spanish, Italian, Portuguese, Korean, Dutch, Chinese, Hindi, and Arabic. However, the training subset does not include tweets written in Arabic, Dutch, Korean, or Hindi. 

The ask can be modeled as a regression task. We explore some of the most successful transformers such as BERT \cite{devlin2019bert}, DistilBERT \cite{sanh2019distilbert}, MiniLM \cite{wang2020minilm}, XLM-RoBERTa \cite{conneau-etal-2020-unsupervised}, and XLM-T\cite{barbieri2022xlm}. In addition, we also study some data augmentation techniques to increase the training dataset. Based on our experiments during the development phase of the competition, we decided to combine the XLM-T transformer model and the data augmentation techniques to estimate our predictions for the test set during the test phase. 

Our team, HULAT, obtained an overall correlation (Pearson's r) of 0.55. The highest correlation was 0.616, while the lowest correlation was 0.03. Our system ranked in the 27th position out of the 45 participating systems. Although our results are modest, our system performs very well for some languages such as Portuguese and English, ranking third and ninth 
in the competition, respectively. As expected, our system obtains worse results for the unseen languages (Arabic, Korean, or Hindi), but it ranks seventh for Dutch. Regarding the data augmentation techniques, our experiments show that the use of synthetic data does not appear to provide a significant improvement in the performance of the transformers. All our code is available in the repository \url{https://github.com/isegura/hulat_intimacy}. 

\section{Background}

%TODO: SUMARIZE THE TASK

This task is a regression task whose goal is to predict a numerical value ranging from 1-5 that represents the intimacy level of a tweet, where 1 is the lowest level, while 5 is the highest level of intimacy. An example is ``Here is a nice equation: 0+0-0-0+0=0", which was annotated with an intimacy score of 1.0. Another example is ``hes eating cake and im planning to eat him", which was annotated with an intimacy score of 4.40.  

The training dataset contains 9,491 tweets from six different languages: Portuguese, Chinese, Spanish, French, English, Italian. The test dataset has 3,881 tweets written in the previous languages, but also in four unseen languages:  Korean, Dutch, Arabic, and Hindi.  Both training and test datasets have balanced representations of their languages (see Fig.  \ref{fig:dist_languages}). In the training dataset, each language is represented by a sample of approximately 1,500 tweets. In the test dataset, the average size is around 400 tweets per language, except for Hindu, which only has 280 tweets. 

We have studied the distribution of the intimacy scores in the training and test datasets (see Fig. \ref{fig:dist_intimacy_scores}). The mean intimacy score is around two tokens in both datasets. 95\% of the tweets have an intimacy score lower than 3.8 in the training dataset, and lower than four in the test dataset. 

Since our dataset is multilingual, we are also interested in studying the distribution of the intimacy scores for each language. In the training dataset, the tweets in all languages tend to have low intimacy scores, with their highest densities at 1.5 score (see Fig. \ref{fig:density-intimacy-language}). English, Italian, and French are the languages that have a larger number of tweets with intimacy scores lower than 2.0. Chinese, Portuguese, and Spanish show more uniform distribution than the others. Chinese and Spanish tweets appear to have the tweets with higher intimacy scores. We also observe the same distribution of the intimacy scores for the six languages in the test dataset (see Fig. \ref{fig:density-intimacy-language-test}).

\begin{figure}[bhp]
\includegraphics[width=\columnwidth]{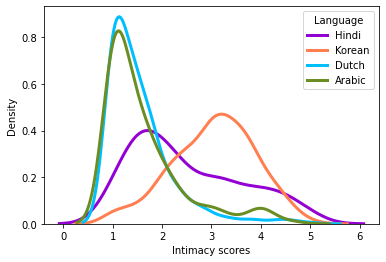}
\caption{Density graph of the intimacy scores  in the test dataset, for the unseen languages: Hindi, Korean, Dutch, and Arabic. }
\label{fig:density_intimacy_language_unseen}
\end{figure}
Regarding the distribution of the intimacy scores for the new languages in the test dataset (that is, Hindi, Korean, Dutch, Arabic), we observe that Dutch and Arabic appear to have a large number of tweets with lower intimacy scores than Korean and Hindi (see Fig. \ref{fig:density_intimacy_language_unseen}). Indeed, 75\% of the tweets written in Dutch or Arabic have intimacy scores lower than 1.8. In both languages, the mean score is around 1.5. On the contrary, 75\% of the tweets written in Hindi have intimacy scores greater than 1.5, with a mean score of 2.5. Korean tweets appear to have even greater intimacy scores, with a mean score of 3.0, and 75\% of them show intimacy scores greater than 2.5. Therefore, textual intimacy may be strongly related to language (see Fig\ref{fig:density-score-chinese-korean}).

\begin{figure}
\includegraphics[width=\columnwidth]{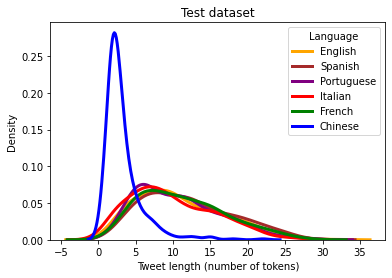}
\caption{Density graph of tweet length in the test dataset for the seen languages: Portuguese, Chinese, Spanish, French, English, Italian.}
\label{fig:histo-length-lang-test}
\end{figure}

We also studied the length of the tweets. The mean size is around 10 tokens in both training and test datasets. Moreover, 95\% of the tweets have less than 22 tokens (see Fig. \ref{fig:histo-length-datasets}).
Figure \ref{fig:histo-length-lang-train} shows the distribution of tweet length for each language in the training dataset. As expected, Chinese tweets are considerably shorter than the other languages, with an average length of between three and four tokens. This is due to the fact that the Chinese language uses ideographs to represent words and ideas, and therefore, these symbols have more information than individual characters. For the romance languages, tweets are larger with an average length of around 10 tokens. All languages show a similar distribution of their lengths. Moreover, the same distribution was seen in the test dataset (see Fig. \ref{fig:histo-length-lang-test}).

\begin{figure}
\includegraphics[width=\columnwidth]{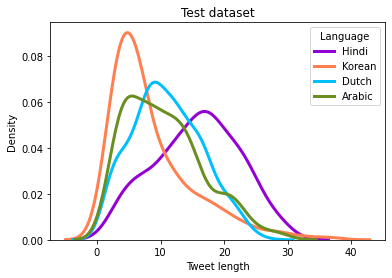}
\caption{Density graph of tweet length in the test dataset for the unseen languages: 'Arabic', 'Dutch', 'Hindi', 'Korean'.}
\label{fig:hist_length_unseen}
\end{figure}

For the unseen languages, Arabic, Dutch, Hindi, and Korean, (which are only present in the test dataset), we can see the tweets written in Korean are shorter than the tweets written in other languages (see Fig. \ref{fig:hist_length_unseen}). This is because Korean is also based on ideographs. In fact, this language is derived from Chinese \cite{goodman2012reading}. Tweets written in Hindi are larger with an average length of 15 tokens approximately. Arabic and Dutch show a similar distribution, with an average length of around 10 tokens.  
%%TODO: comentar los lenguajes no vistos con más detalle.Investigar como es el HIndi, cómo es el ARabic y el Dutch.  

\section{System Overview}
\label{sec:systems}
\subsection{Models}

We now provide a brief description of the transformers that we explored during the development phase. 

BERT \cite{devlin2019bert} is undoubtedly the most popular transformer model due to its excellent results in many NLP tasks. In fact, many of the transformers that have been published since then (such as DistilBERT, RoBERTa or XLM-R) are based on BERT. BERT is an encoder trained using two strategies: masked language modeling (MLM) and next sentence prediction (NSP). The multilingual version of BERT was pre-trained in more than one hundred languages using Wikipedia. DistilBERT \cite{sanh2019distilbert} and MiniLM \cite{wang2020minilm} are smaller versions of BERT, which can achieve similar results to BERT but with less training time.

XLM-RoBERTa (XLM-R) \cite{conneau-etal-2020-unsupervised} was pre-trained using more than two terabytes of texts from the Common Crawl Corpus \footnote{\url{https://commoncrawl.org/the-data/}}, which was collected by using web crawling for 12 years. Like BERT, XLM-R is based on the strategy of masked language modeling. This is a multilingual model that was pre-trained in one hundred languages. However, XLM-R still provides excellent results for monolingual text classification. Moreover, it also obtains good performance for low-resource languages, such as Swahili and Urdu. XLM-T \cite{barbieri2022xlm}, based on XLM-R, was pre-trained on millions of tweets in over thirty languages.

\subsection{Data augmentation}

Data augmentation (DA) aims to increase the training size by applying different transformations to the original dataset. For example, in computer vision, some modifications can be performed by cropping, flipping, changing colors, and rotating pictures. In NLP, these transformations include swapping tokens (but also characters or sentences), deletion or random insertion of tokens (but also characters or sentences), and back translation of texts between different languages. While those transformations are easier to implement in computer vision, they are challenging in NLP, because they can alter the grammatical structure of a text.

Another advantage is that these techniques help to enhance the diversity of the examples in the dataset. Moreover, they also help to avoid overfitting. 
Unfortunately, data augmentation does not always improve the results in NLP tasks.

In this task, we used different data augmentation techniques (such as EDA \cite{wei-zou-2019-eda}, and NLPAug library\footnote{\url{https://github.com/makcedward/nlpaug}}) to create synthetic data.

EDA has been implemented in the textaugment library  \footnote{\url{https://github.com/dsfsi/textaugment}} for Python. EDA uses four simple operations:  Synonym Replacement, Random Insertion, Random swap, and Random Deletion. 
The first operation randomly chooses n words in a sentence (which are not stopwords). Then, these words are replaced with synonyms from WordNet\footnote{\url{https://wordnet.princeton.edu/}}, a very large lexicon for English. 
Random insertion chooses a random word (which is not a stopword). Then, it finds a random synonym that is inserted in a random position in the sentence. The third operation, Random Swap,  randomly chooses two words in the sentence and swaps their positions. The fourth operation, Random Deletion, randomly removes a word from a sentence. These operations can be repeated several times. 

NLPAug also provides an efficient implementation of DA techniques. In particular, NLPAug offers three types of augmentation: Character level augmentation, Word level augmentation, and 
Sentence level augmentation. In each of these levels, NLPAug provides all the operations described above, that is, synonym replacement, random deletion, random insertion, and swapping. Regarding  synonym replacement, the most effective way is using word embeddings to select the synonyms. This technique allows us to obtain a sentence with the same meaning but with different words. NLPAug uses non-contextual embeddings (such as Glove, word2vec, etc) or contextual embeddings (such as Bert, Roberta, etc).

In this work, we use the synonym replacement provided by EDA, which is based on WordNet. Thanks to NLPAug, we also generate new examples by using a contextualized language model such as BERT.

\section{Experimental Setup}
During the development phase, we divided the training dataset into three splits with a ratio of 70:10:20. That is,  6,643 texts for training, 940 for validation, and 1,908 for testing. These splits allowed us to train and  validate our models, and choose the best model for the test phase. The three splits have the same distribution of languages and intimacy scores. 

Before tokenization, we used some filters to clean texts. For example, we removed all mentions of '@user' and URLs. We also removed punctuation. Based on the length distribution of texts (see Fig. \ref{fig:histo-length-datasets}), we consider 50 tokens as the maximum length. All models were trained with batch size 64 for three epochs, while the batch size was 20 for validation. We train all models using Pearson's r as the metric to choose the best model. Pearson correlation coefficient measures the linear relationship between two datasets, in our case, the actual and predicted values. It can range from -1 to +1, where 0 means that there is no correlation. A positive score tells us that there is a positive association between the two datasets.

\section{Results}
Our team, HULAT, participated in the task by using the XLM-T model \cite{barbieri2022xlm} and data augmentation. Our approach obtained an overall correlation (Pearsons'r) of 0.55. The highest correlation was 0.616, while the lowest correlation was 0.03. Our system ranked in the 27th position out of the 45 participating systems. 

Regarding the results for the seen languages (that is, the languages that are in the training dataset), our system ranked third for Portuguese with a correlation of 0.699, very close to the highest correlation (0.702). For English, our system ranked ninth for English with a correlation of 0.721, being the highest correlation of 0.758.  Table \ref{tab:results_seen_languages} shows our results on the test dataset for these languages. Overall, our results are five points below the best score, except for Spanish, where the difference is even higher (around eight points).

\begin{table}
  \centering
  \small
  \begin{tabular}{lccc}
  \hline
  \textbf{Language} &  \bf Pearson & \bf Top Pearson & \bf Ranking\\
  \hline
  Portuguese & .699&.702& 3\\
  English &.721 &	.758& 9\\
  Chinese & .709& .762 & 21\\
  Italian &.693 &	.742 & 22 \\
  French & .670 & 726 & 25\\
  Spanish &.698 &	.784 & 29 \\
  Overall & .707 & .707 & 18\\
  \hline
  \end{tabular}
  \caption{XLM-T results on the test dataset  for the seen languages. The third column shows the highest correlation in the competition for each language. The fourth column shows our ranking in the competition for each language.}
  \label{tab:results_seen_languages}
\end{table}

Regarding the results for the unseen languages (that is, the languages that were not in the training dataset), our system achieved a correlation of 0.355, while the highest correlation was 0.449. However, our system ranked eighth for Dutch with a correlation of 0.642 (the highest correlation was 0.678).
Table \ref{tab:results_unseen_languages} shows our results on the test dataset for each of the unseen languages.
There is still a lot to improve, especially in Korean (our score is almost 17 points below the highest correlation).

\begin{table}
  \centering
  \small
  \begin{tabular}{lccc}
    \hline

  \textbf{Language} &  \bf Pearson &   \bf Top Pearson & \bf Ranking\\
  \hline
  Dutch & .641 & .678 &7\\
  Arabic & .601 &	.662 & 22 \\
  Hindi & .209& .276 & 25\\
 Korean & .256 &	.419& 35 \\
  Overall & .355 & .499 &33\\
  \hline
  \end{tabular}
  \caption{XLM-T results for unseen languages on the test dataset. The third column shows the highest correlation in the competition for each language. The fourth column shows our ranking in the competition for each language.}
  \label{tab:results_unseen_languages}
\end{table}

During the development phase, we explored different models described in Section \ref{sec:systems}. Table \ref{tab:pearson_scores} shows the results (Pearson's r) of all our models on the final test dataset. The best model is XLM-T. This may be because this model was pre-trained on millions of tweets in over thirty languages. The second model was XLM-R. Both models, XLM-R and XLM-T, are multilingual and have the same architecture. The only difference is that the first one was pre-trained with texts from the Common Crawl Corpus, while the second one was pre-trained with tweets. This fact could explain why XLM-T outperforms XLM-R. 
DistilBERT is the third model, achieving better results than BERT.  Like DistilBERT, MiniLM is also a smaller version of BERT, but with worse results than DistilBERT. 

\begin{table}
  \centering
  \small
  \begin{tabular}{llc}
  \textbf{Model} & \textbf{Aug.} & \bf Pearson\\
  \hline
  
\multirow{2}{*}{MiniLM} & No &.457\\
                  & Yes & .501\\
                  
  \multirow{2}{*}{BERT} & No &.475\\
                  & Yes &.468\\

  \multirow{2}{*}{DistilBERT} & No &.52\\
                  &Yes &.489\\
  
\multirow{2}{*}{XLM-R} & No &.526\\
                  & Yes &.527\\

\multirow{2}{*}{XLM-T} & No &.565\\
                  & Yes &.563\\

\hline
  \end{tabular}
  \caption{Pearson scores for all our models on the final test dataset}
  \label{tab:pearson_scores}
\end{table}

Data augmentation does not appear to affect the results of the XLM-R and XLMT models. For BERT and DistilBERT, the use of synthetic data even decreases their results. MiniLM is the only model that appears to benefit from data augmentation techniques.

\section{Conclusion}

In this paper, we fine-tuned some of the most successful transformers by using an augmented training dataset created with some data augmentation techniques. Based on the results during the development phase, we chose the XLM-T model with data augmentation for the test phase. Our system obtained an overall correlation of 0.55, six points below the highest correlation in the ranking. For Portuguese and English, it ranked fourth and ninth, respectively. Although our results for unseen languages are modest, our system ranked eight for Dutch.  

As future work, we plan to use zero-shot text classification methods to automatically annotate a small sample of tweets written in the unseen languages of the task, which later will be manually reviewed by experts. We also plan to extend the dataset by using texts from novels (which should contain more intimacy) and from academic books (whose texts, in general, should have a less degree of intimacy). Fortunately,  novels and academic books are usually published in different languages. A group of experts will annotate a small sample of these texts. Then, we will explore some data augmentation techniques, especially based on back translation, to create new synthetic data for the task and for all the languages.

\section{Acknowledgments}
This work is part of the R\&D\&i ACCESS2MEET project (PID2020-116527RB-I0), financed by MCIN AEI/10.13039/501100011033/. 
%This work was also supported  by the line of Excellence of University Professors (EPUC3M17). 

%\bibliography{custom}
%\bibliographystyle{acl_natbib}

\section{Appendix}
\label{sec:appendix}

\begin{figure}[bhp]
\includegraphics[width=\columnwidth]{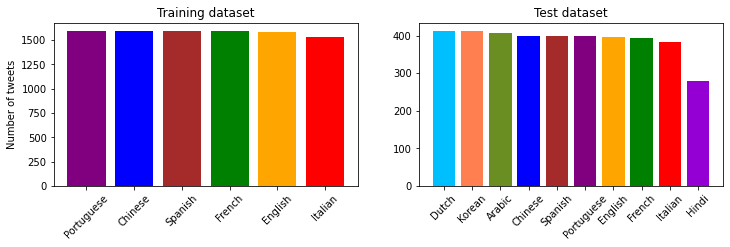}
\caption{Distribution of the languages in the training and test datasets. }
\label{fig:dist_languages}
\end{figure}

\begin{figure}[bhp]
\includegraphics[width=\columnwidth]{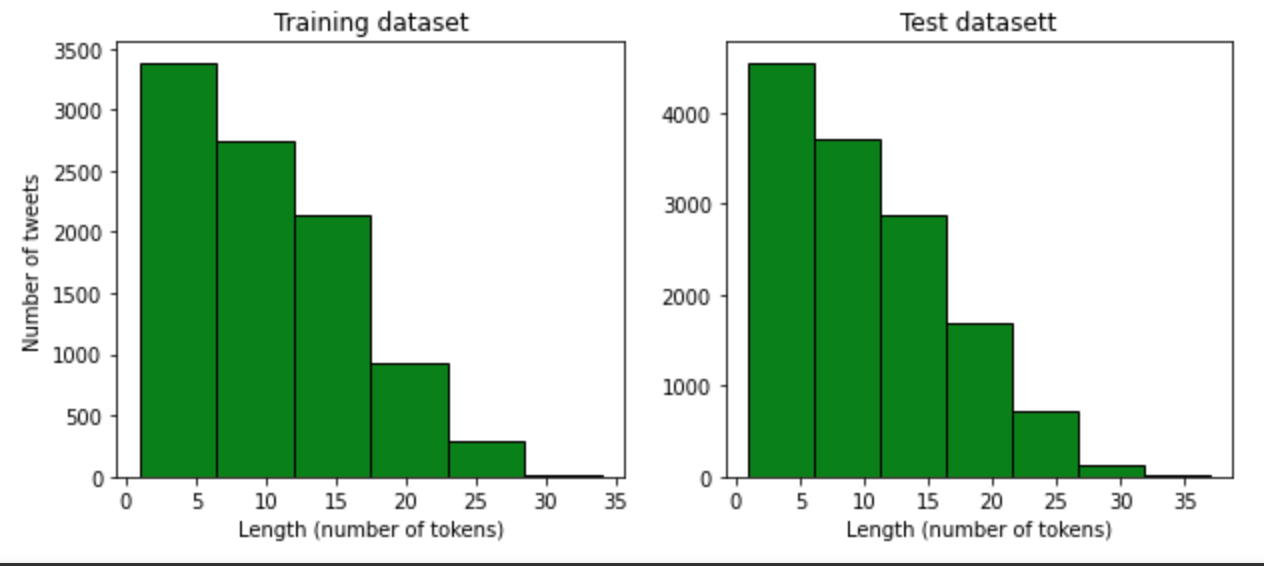}
\caption{Distribution of the length of texts in tokens.}
\label{fig:histo-length-datasets}
\end{figure}

\begin{figure}
\includegraphics[width=\columnwidth]{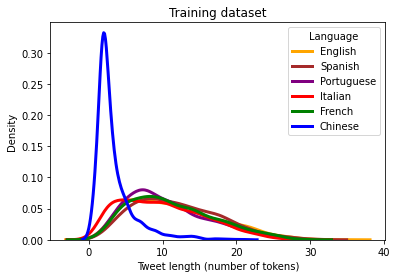}
\caption{Density of tweet length in the training dataset.}
\label{fig:histo-length-lang-train}
\end{figure}

In this section, we provide supplementary material for our research. 
Figure \ref{fig:dist_languages} shows the distribution of languages in the training and test datasets. Figure \ref{fig:histo-length-datasets} is a histogram of the length of the tweets in both datasets. Figure \ref{fig:histo-length-lang-train} shows a density graph of the length of the tweets in the training dataset.  

\begin{figure}
\includegraphics[width=\columnwidth]{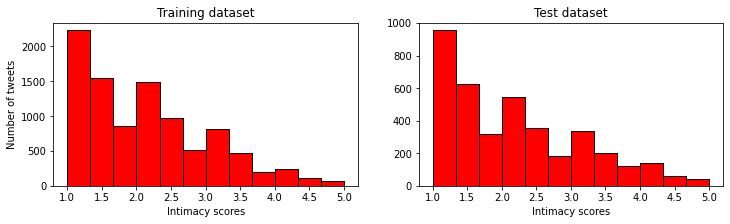}
\caption{Histogram of the intimacy scores in the training and test datasets. }
\label{fig:dist_intimacy_scores}
\end{figure}

\begin{figure}
\includegraphics[width=\columnwidth]{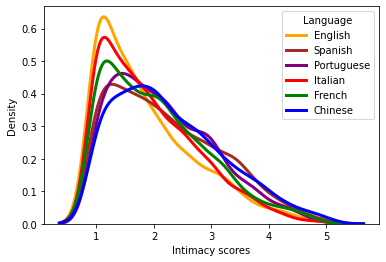}
\caption{Density graph of the intimacy scores in the training dataset}
\label{fig:density-intimacy-language}
\end{figure}

\begin{figure}[bhp]
\includegraphics[width=\columnwidth]{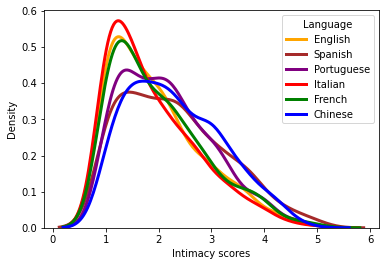}
\caption{Density graph of the intimacy scores in the test dataset for the seen languages: Portuguese, Chinese, Spanish, French, English, Italian}
\label{fig:density-intimacy-language-test}
\end{figure}

\begin{figure}
\includegraphics[width=\columnwidth]{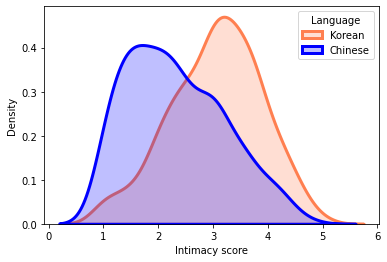}
\caption{Comparison of the intimacy scores for Chinese and Korean in the test dataset. }
\label{fig:density-score-chinese-korean}
\end{figure}

Figure \ref{fig:dist_intimacy_scores} is a histogram of the intimacy scores in training and test datasets.  Figures \ref{fig:density-intimacy-language} and \ref{fig:density-intimacy-language-test}  show the density graph of the intimacy scores for each language in the training and test dataset, respectively. 
Figure \ref{fig:density-score-chinese-korean} shows a comparison of the intimacy scores for Chinese and Korean in the test dataset, which shows that textual intimacy may be strongly related to language.

During training, we also compute some common metrics for regression tasks (see Table \ref{tab:results}), which are described below: 
\begin{itemize}
\item Mean Absolute Error(MAE) calculates the absolute difference between actual and predicted values. That is, we must sum all the errors and divide them by the total number of instances in the test dataset. 
\item Mean Squared Error(MSE) is the squared distance between actual and predicted values. 
\item Root Mean Squared Error(RMSE) is the root of MSE. 
\item R2 score (also known as Coefficient of Determination) ranges from 0 to 100. A low value tells us that our regression model is not working well. Unlike the previous metrics, this gives us the performance of the model, not the error. 
\end{itemize}

\begin{table}[hbp]
  \centering
  \tiny
  \begin{tabular}{llccccc}
  \textbf{Model} & \textbf{Aug.} & \bf MSE & \bf RMSE & \bf MAE & \bf SMAPE & \bf R2 \\
  \hline
  
\multirow{2}{*}{MiniLM} & No & .771&.878&.667&31.2&.171\\
                  & Yes & .704&.839&.650&30.7&.243\\
                  
  \multirow{2}{*}{BERT} & No & .762&.873&.661&30.9&.18\\
                  & Yes & .797&.892&.657&30.9&.14\\

  \multirow{2}{*}{DistilBERT} & No & .693&.832&.635&29.8&.255\\
                  &Yes & .764&.874&.65&30.5&.179\\
  
\multirow{2}{*}{XLM-R} & No & .708&.841&.640&30.1&.238\\
                  & Yes & .747&.864&.653&30.2&.196\\

\multirow{2}{*}{XLM-T} & No & .648&.805&.625&29.8&.304\\
                  & Yes & .645&.803&.615&29.1&.306\\

\hline
  \end{tabular}
  \caption{Results on the final test dataset for the metrics: MSE,  RMSE,  MAE, SMAPE, and R2  }
  \label{tab:results}
\end{table}
\end{document}